%% file: main.tex
\pdfoutput=1
\documentclass{article}

\usepackage[final, nonatbib]{neurips_2023}

\usepackage[utf8]{inputenc} % allow utf-8 input
\usepackage[T1]{fontenc}    % use 8-bit T1 fonts
\usepackage{hyperref}       % hyperlinks
\usepackage{url}            % simple URL typesetting
\usepackage{booktabs}       % professional-quality tables
\usepackage{amsfonts}       % blackboard math symbols
\usepackage{nicefrac}       % compact symbols for 1/2, etc.
\usepackage{microtype}      % microtypography
\usepackage{xcolor}         % colors

\usepackage[backend=biber,style=ieee, citestyle=numeric-comp, sortcites]{biblatex}

\DeclareSortingTemplate{nonenyt}{
  \sort{\citeorder}
  \sort{
    \field{presort}
  }
  \sort[final]{
    \field{sortkey}
  }
  \sort{
    \field{sortname}
    \field{author}
    \field{editor}
    \field{translator}
    \field{sorttitle}
    \field{title}
  }
  \sort{
    \field{sortyear}
    \field{year}
  }
  \sort{
    \field{sorttitle}
    \field{title}
  }
  \sort{
    \field{volume}
    \literal{0}
  }
}
\usepackage{biblatex2bibitem}
\input{000.preamble}

\input{010.title}

\begin{document}
\maketitle
\input{099.abstract}
\input{100.introduction}

\input{300.approach}
\input{400.experiments}

\input{200.background}

\input{500.conclusion}

% \newpage
\input{600.ethics}

\newpage
\printbibliography
% \bibliography{references}

\newpage
\appendix
\input{a00.example}

\end{document}

%% file: 000.preamble.tex
\usepackage{times}
\usepackage{latexsym}

\usepackage[T1]{fontenc}
\usepackage{tabularray}
\usepackage{arydshln}
\usepackage{scrextend}
\usepackage[utf8]{inputenc}

\usepackage{inconsolata}

\usepackage{amsmath}
\usepackage{amssymb}
\usepackage{array}
\usepackage{booktabs}
\usepackage[font=small]{caption}
\usepackage{color}
\usepackage{colortbl}
\usepackage{comment}
\usepackage{enumitem}
\usepackage{epsfig}
\usepackage{extarrows}
\usepackage{fontawesome}
\usepackage[T1]{fontenc}
\usepackage{graphicx}
\usepackage{hyperref}
\usepackage[misc]{ifsym}
\usepackage{makecell}
\usepackage{multirow}
\usepackage{soul}
\usepackage{tabularx}
\usepackage{xcolor}
\usepackage[export]{adjustbox}
\usepackage{xifthen}
\usepackage{xspace}

\usepackage{tabularray}
\usepackage{makecell}
\usepackage{booktabs}
\usepackage{multirow}
\usepackage{extarrows}

\usepackage{amssymb}% http://ctan.org/pkg/amssymb
\usepackage{pifont}% http://ctan.org/pkg/pifont

\newcommand{\cmark}{\ding{51}}%
\newcommand{\xmark}{\ding{55}}%

\usepackage{microtype}
\usepackage{todonotes}

\usepackage{xcolor}

\newcommand{\ourFramework}{SpeechVerse\xspace}

%% file: 010.title.tex
\title{\ourFramework: A Large-scale Generalizable Audio Language Model}

\author{
Nilaksh Das $^*$ \And Saket Dingliwal \thanks{Equal Contributions}  \thanks{Corresponding author: skdin@amazon.com} \And Srikanth Ronanki    \And Rohit Paturi    \And Zhaocheng Huang    \And Prashant Mathur    \And  Jie Yuan    \And  Dhanush Bekal    \And  Xing Niu    \And  Sai Muralidhar Jayanthi    \And  Xilai Li    \And  Karel Mundnich     \And  Monica Sunkara    \And  Sravan Bodapati \And Sundararajan Srinivasan    \And  Daniel Garcia-Romero    \And  Kyu J. Han    \And Katrin Kirchhoff   \\ \\ \\ \large \textbf{AWS AI Labs, Amazon}}

%% file: 099.abstract.tex
\begin{abstract}
%Large language models (LLMs) have demonstrated incredible proficiency in performing tasks that require semantic understanding of natural language instructions.
%
%Recently, many works have further expanded this capability to perceive multimodal inputs, thus enabling LLMs to ``see'' and ``listen''.
%
%However, such multimodal perception is often confined in scope to the limited tasks on which an LLM is finetuned.
%
%Moreover, for spoken language understanding (SLU) systems, a bigger challenge emerges in learning latent representations for acoustic variability (e.g., from different accents, pronunciations etc.) that may be objectively optimal for a wide variety of SLU tasks.
%
%In this work, we develop a robust framework, \ourFramework, to address these challenges.
%
%In our approach, we propose a novel method that simultaneously induces speech features at varying levels of sampling resolution.
%
%This allows the attention mechanism of the model  to systematically learn composite hierarchies  of representations optimal for diverse SLU tasks.
%
%Further, we improve the zero-shot capability of the model to infer unseen class labels by successively eliciting compound goals in a manner that maximizes introspection from  the autoregressively produced output.
%
%Our extensive empirical experiments reveal that \ourFramework improves zero-shot performance for \fixme{task TBD} by \fixme{xxx\%} when compared to a competitive baseline.

%Experiments are conducted on publicly available speech corpora and 

Large language models (LLMs) have shown incredible proficiency in performing tasks that require semantic understanding of natural language instructions. Recently, many works have further expanded this capability to perceive multimodal audio and text inputs, but their capabilities are often limited to specific fine-tuned tasks such as automatic speech recognition and translation. We therefore develop SpeechVerse, a robust multi-task training and curriculum learning framework that combines pre-trained speech and text foundation models via a small set of learnable parameters, while keeping the pre-trained models frozen during training. The models are instruction finetuned using continuous latent representations extracted from the speech foundation model to achieve optimal zero-shot performance on a diverse range of speech processing tasks using natural language instructions. We perform extensive benchmarking that includes comparing our model performance against traditional baselines across several datasets and tasks. Furthermore, we evaluate the model's capability for generalized instruction following by testing on out-of-domain datasets, novel prompts, and unseen tasks. Our empirical experiments reveal that our multi-task SpeechVerse model is even superior to conventional task-specific baselines on 9 out of the 11 tasks. 

%, demonstrating its formidable instruction following skills.

%which takes natural language instructions as input to perform various speech processing tasks through supervised instruction finetuning. Our approach

\end{abstract}

%% file: 100.introduction.tex
\section{Introduction}
\label{sec:introduction}
Large language models (LLMs) \cite{radford2018improving, brown2020language, chowdhery2023palm} have achieved remarkable performance on a variety of natural language tasks through self-supervised pre-training on massive text corpora. They have also shown a striking ability to follow open-ended instructions from users through further instruction tuning \cite{chung2022scaling, ouyang2022training, achiam2023gpt, touvron2023llama}, enabling strong generalization capabilities. Despite the success, a significant limitation lies in the inability of language models to perceive non-textual modalities such as images and audio. % and is critical for progress towards artificial general intelligence (AGI) \cite{goertzel2007artificial, goertzel2014artificial}.

Speech in particular represents the most natural mode of human communication. Empowering LLMs to deeply understand speech could significantly enhance human-computer interaction \cite{huang2023audiogpt} and multimodal dialog agents \cite{team2023gemini, guo2024large}. As such, enabling LLMs to comprehend speech has received substantial attention recently. Some approaches first transcribe speech via an automated speech recognition (ASR) system and then process the text with an LLM for improved transcription \cite{li2023prompting, ma2023can, huang2024multilingual}. However, such pipelines cannot capture non-textual paralinguistic and prosodic features like speaker tone, intonation, emotion, valence, etc. 
%additional encoders that can process speech and other audio signals
%Another emerging research direction focuses on directly fusing textual LLMs with speech encoders, allowing end-to-end training to integrate multimodal signals beyond text transcription, for richer speech and audio understanding \cite{wang2023viola, rubenstein2023audiopalm}. 

\begin{figure*}[t]
    \centering
      \includegraphics[width=0.9\linewidth]{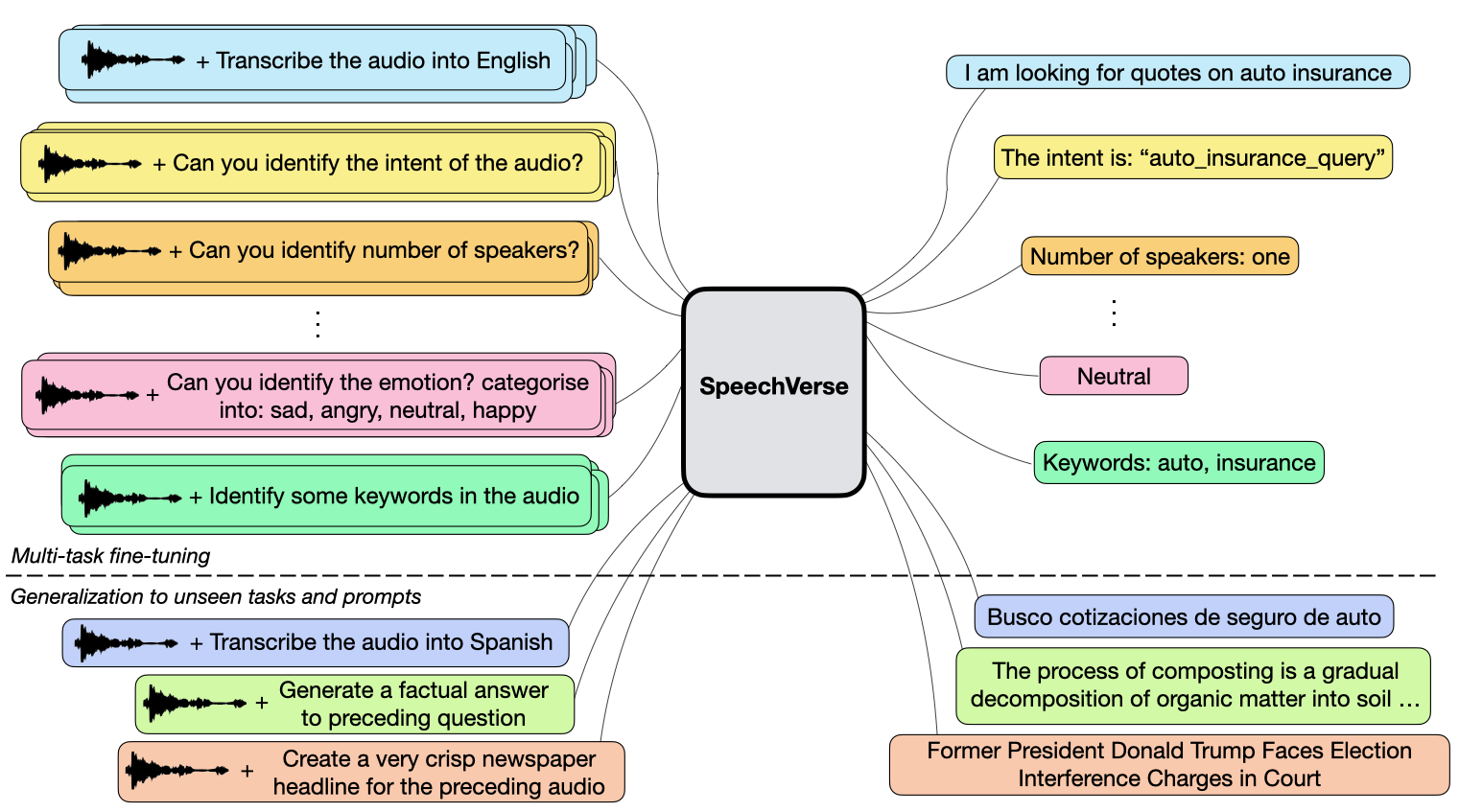}
      %\vspace{2mm}
      \caption{Schematic diagram of the SpeechVerse framework.}
      \label{fig:speechverse_schema}
      \vspace{-4mm}
\end{figure*}

A promising new paradigm directly fuses textual LLMs with speech encoders within an end-to-end training framework \cite{wang2023viola, rubenstein2023audiopalm}. Enabling joint modeling of speech and text holds promise for richer speech and audio comprehension versus text-only methods. Particularly, instruction-following multimodal audio-language models \cite{zhang2023speechgpt, wang2023slm, chu2023qwen} are increasingly receiving more attention due to their generalization capability. Despite some success, existing works in multi-task audio-language models such as SpeechT5 \cite{ao2021speecht5}, Whisper \cite{radford2023robust}, VIOLA \cite{wang2023viola}, SpeechGPT \cite{zhang2023speechgpt}, and SLM \cite{wang2023slm} are limited to processing a small number of speech tasks. 

%More recently, \citet{chu2023qwen} proposed Qwen-Audio, a multi-task audio-language model capable of perceiving human speech and sound signals and is trained on 30 tasks from diverse audio types including music and songs. However, this requires carefully designed hierarchical tagging and a large-scale supervised audio encoder for fusion making it suboptimal for diverse speech tasks. 

We therefore propose SpeechVerse, a robust multi-task framework that leverages supervised instruction finetuning to incorporate various speech tasks (see Figure \ref{fig:speechverse_schema}). In contrast to SpeechGPT \cite{zhang2023speechgpt}, we propose using continuous representations extracted from a self-supervised pre-trained speech foundation model, focusing on tasks that generate text-only output. More recently, \cite{chu2023qwen} proposed Qwen-Audio, a multi-task audio-language model capable of perceiving human speech and sound signals and is trained on 30 tasks from diverse audio types including music and songs. However, this requires carefully designed hierarchical tagging and a large-scale supervised audio encoder for fusion making it sub-optimal for unseen speech tasks.  In contrast, our training paradigm incorporates multi-task learning and supervised instruction finetuning in a unified curriculum, without the need for task-specific tagging, allowing for generalization to unseen tasks using natural language instructions.
% \cite{rubenstein2023audiopalm, chu2023qwen}. 

%\pra{Prashant: At a high level, main differentiator is ability to scale via SIFT. I would remove the first differentiator.} 

\noindent
% To summarise, we make the following novel contributions
% through this work:
%\textbf{Contributions}
\noindent We summarize our contributions below:
\begin{enumerate}[noitemsep, nosep, left=0pt]
    \item \textbf{Scalable multimodal instruction finetuning for diverse speech tasks.} 
    SpeechVerse is a novel LLM-based audio-language framework to scalably exhibit strong performance on as many as 11 diverse tasks. We
    extensively benchmark our models  on public datasets spanning ASR, spoken language understanding and paralinguistic speech tasks.
    
    \item \textbf{Versatile instruction following capability for novel open-ended tasks.} 
    We demonstrate the SpeechVerse model's capability to leverage 
    the robust language understanding of the LLM backbone in order to adapt to
    open-ended tasks that were unseen during multimodal finetuning.

    \item \textbf{Strategies for improving generalization to unseen tasks.}
    We further study prompting and decoding strategies including constrained and joint decoding,
    that can enhance the model's ability to generalize to completely unseen tasks, 
    improving absolute metrics by up to 21\%.
\end{enumerate}

%% file: 300.approach.tex
\section{Approach}
\label{sec:approach}
\subsection{Architecture}
\begin{figure}[t]
    \centering
      \includegraphics[scale=0.3]{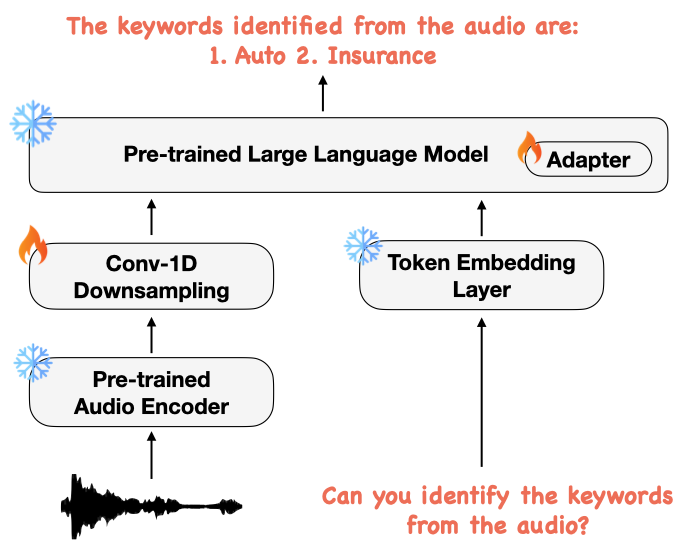}
      \caption{Block diagram of the SpeechVerse architecture.}
      % \vspace{-2mm}
      \label{fig:speechverse_arch}
      % \vspace{-5mm}
\end{figure}
As shown in Figure \ref{fig:speechverse_arch}, our multimodal model architecture consists of three main components:
(1) a pre-trained audio encoder to encode an audio signal into a feature sequence,
(2) a 1-D convolution module that operates over the audio feature sequence to abbreviate the sequence length,
and (3) a pre-trained LLM to use these audio features and textual instructions to perform the required task. 
Details of each of these sub-systems are described below.

\smallskip

\noindent\textbf{Audio encoder:}
To extract semantic features from a given audio,
we use a large pre-trained self-supervised speech foundation model 
% such as WavLM \cite{Chen2021WavLMLS}
as an audio encoder. 
We can represent the audio encoder as a cascading collection
of $L$ layers, where each intermediate layer $l$
returns a feature sequence
$\mathbf{h}^{(l)} = f_{AE}(\mathbf{h}^{(l-1)}; \theta^{(l)}_{AE})$,
where $\mathbf{h}^{(0)} = \mathbf{x}$ is the input audio.
Here, $\theta^{(l)}_{AE}$ represents the learned weights
for layer $l$ of the pre-trained speech model.
%
% In all our experiments, we keep the pre-trained model weights $\theta^{(l)}_{AE}$ frozen.
%
To capture a unified representation for
multi-faceted forms of feature semantics,
we compute the output of the audio encoder as:
\vspace{-0.2cm}
\begin{align}
% \resizebox{0.6\linewidth}{!}{
    AE(\mathbf{x}) = \frac{1}{L} \sum_{l=1}^L w^{(l)} \mathbf{h}^{(l)}
% }
\vspace{-0.2cm}
\end{align}
where, the scalars $\{w^{(1)}, \dots w^{(L)}\}$ 
are a set of learnable parameters.
%
% This approach allows us to simultaneously learn a diverse set of tasks
% by capturing learned features from different intermediate layers
% of the speech foundation model, as different forms of semantics 
% may be important for different tasks.
As this approach concurrently encodes features 
from multiple intermediate layers of the speech foundation model,
it can simultaneously capture different forms of semantics
(higher-order as well as lower-order features),
leading to better generalization across a diverse set of tasks.
We also perform experiments by only taking the output 
from the last layer of the audio encoder, i.e.,
$AE(\mathbf{x}) = \mathbf{h}^{(L)}$.

\smallskip

\noindent\textbf{Convolution downsampling module:}
LLMs trained on text-only input 
encode token sequences
which are typically of a considerably shorter length
compared to the feature sequences encoded by
speech foundation models.
To mitigate this large discrepancy in length distribution
between audio features and text tokens,
we downsample the encoded audio features through 
a learnable convolution module.
This module consists of successive blocks,
each having a 1-D convolution layer 
followed by layer normalization.
For the 1-D convolution,
we use a kernel size of 3,
% with a stride of 2.
%
% This leads to a 2x downsampling (or halving) 
% of the sequence length
% with each successive block.
% The kernel size of 3 
which ensures that each output frame
has both left and right context from the input frames.
In our experiments, we use as many number of these downsampling blocks
as necessary such that the resulting
sampling rate for the audio comes out to be 12.5 Hz, i.e.,
each output frame corresponds to 80ms of audio.
By fine-tuning the convolution downsampling module,
the audio encoder output can be transformed
from an audio-only feature space
to a joint audio-text semantic space.
Hence, we set the number of output channels for the 1-D convolution
to be equal to the feature dimension of the token embeddings
for the downstream LLM.
We can represent the output of the convolution downsampling module as
$CNN(\mathbf{x}) = f_{CNN}(AE(\mathbf{x}); \theta_{CNN})$,
where $\theta_{CNN}$ represents learnable parameters of the 1-D convolution blocks.

\smallskip

\noindent\textbf{Large Language Model:} 
An LLM typically takes in a sequence of input text tokens $\mathbf{z}$ 
and models the probability of observing an output text sequence $\mathbf{y}$ 
as the possible next tokens for the input text. 
The text tokens are converted to vectorized embeddings $EMB(\mathbf{z})$ 
using a lookup matrix that is learned during training. 
The output of the LLM can be represented as
$LLM(\mathbf{z}) = f_{LLM}(EMB(\mathbf{z}); \theta_{LLM})$,
where $\theta_{LLM}$ are the weights of the LLM.
In this work, we leverage a pre-trained LLM for multimodal tasks.
% We keep the pre-trained LLM weights $\theta_{LLM}$ frozen,
% and adopt parameter-efficient finetuning techniques for learning
% the speech-based tasks.
%
For formulating the multimodal input with audio $\mathbf{x}$ and text sequence $\mathbf{z}$, we simply concatenate
the downsampled audio features ($CNN(\mathbf{x})$) with the token embeddings ($(EMB(\mathbf{z})$)
in the sequence dimension as shown in Figure \ref{fig:speechverse_arch}.
Hence, we can represent the probability distribution over the output from our multi-modal Spoken Language Model (SLM) as:
% \vspace{-0.2cm}
\begin{align}
    SLM(\mathbf{x}, \mathbf{z}) 
    = f_{LLM}\Big(\big[CNN(\mathbf{x}), EMB(\mathbf{z})\big]; \theta_{LLM}\Big)
    % = f_{LLM}(CNN(\mathbf{x}) \oplus EMB(\mathbf{z}); \theta_{LLM})
% \vspace{-0.2cm}
\end{align}

\subsection{Multimodal Instruction Finetuning}
% \noindent\textbf{Multi-modal Speechverse Model}  As showcased in Figure \ref{fig:arch}, our joint model concatenates the audio feature sequence to the text embeddings sequence and then they are passed through the LLM.  
% \xing{We can introduce the prompt set $\mathcal{P}^\tau$  more naturally. We start with $\mathcal{D}^\tau = \{\mathbf{x}^\tau, \mathbf{p}^\tau ,\mathbf{y}^\tau\}_{1}^{n^\tau}$, where $\mathbf{p}^\tau$ is traditionally a fixed prompt/instruction per task. We extend it to a set of prompts/instructions per task for reasons XYZ. Do we have ablation/justification of doing this?}
Let $\mathcal{D}^\tau = \{\mathbf{x}^\tau, \mathbf{y}^\tau\}_{1}^{n^\tau}$ represent a labelled dataset for a task $\tau$ with $n^\tau$ samples where each sample consists of audio sequence $\mathbf{x}^\tau$ and a corresponding text label sequence $\mathbf{y}^\tau$. 
Let $\mathcal{P}^\tau = \{\mathbf{p}^\tau\}_{1}^{m^\tau}$ be a set of $m^\tau$ textual prompts/instructions sequences $\mathbf{p}^\tau$ describing the task.
In our experiments, we do a weighted combination of datasets from each of the $M$ training tasks $\{w^{\tau},\mathcal{D}^{\tau}\}_1^M$ where $w^{\tau}$ is the weight given to each sample of task $\tau$. This is done to ensure balance between tasks with different complexities and training data sizes. Thus, each sample can be represented as a tuple $(\mathbf{x}^\tau, \mathbf{p}^\tau ,\mathbf{y}^\tau)$ where $\mathbf{x}^\tau$ is the audio sample, $\mathbf{p}^\tau$ is a prompt/instruction sampled uniformly from $\mathcal{P}^\tau$ and $\mathbf{y}^\tau$ is the label. 
Then the probability for predicting the labels $\mathbf{y}^\tau$ can be defined as: 
% \{\theta_{AE}, \theta_{CNN}, \theta_{E}, \theta_{LLM} \}
% \vspace{-0.2cm}
\begin{align}
    p(\mathbf{y}^\tau |\mathbf{x}^\tau, \mathbf{p}^\tau ; \Theta)
    = SLM(\mathbf{x}^\tau, \mathbf{p}^\tau)
    % = ALM( CNN_{\theta_{CNN}}(AE_{\theta_{AE}}(\mathbf{x^\tau})) \oplus E_{\theta_{E}}(\mathbf{p^\tau}), y)
% \vspace{-0.2cm}
\end{align}
where $\Theta = \{\theta_{AE}, \theta_{CNN}, \theta_{LLM} \}$ are all the parameters of our audio language model.
The self-attention layers in the LLM attend to both the audio and the textual instruction to generate the required output on any audio-related task.
We use the standard gradient descent method to maximise the likelihood of producing the target label $\mathbf{y}^\tau$ for each sample in the training dataset as defined below:
% \vspace{-0.2cm}
\begin{align}
 \mathcal{L} (\Theta) = - \log p_{(\mathbf{x}^\tau, \mathbf{p}^\tau ,\mathbf{y}^\tau) \sim \{\mathcal{D}^{\tau}\}_1^M} (\mathbf{y}^\tau |\mathbf{x}^\tau, \mathbf{p}^\tau; \Theta)
% \vspace{-0.2cm}
\end{align}
    % \mathcal{L} (\Theta) = - \log p_{(\mathbf{x}^\tau, \mathbf{p}^\tau ,\mathbf{y}^\tau) \sim \{\mathcal{D}^{\tau}\}_1^M} (\mathbf{y}^\tau |\mathbf{x}^\tau, \mathbf{p}^\tau; \Theta) 
% \end{equation}

\subsection{Curriculum Learning with Parameter Efficient Finetuning}
\label{sec:training}
To ensure faster convergence and avoid catastrophic forgetting and overfitting
by the pre-trained LLM, 
we adopt a parameter efficient approach
based on low-rank adaptation \cite{hu2021lora}, or LoRA,
for training our multimodal models.
In this work, we freeze the pre-trained audio encoder and LLM,
and only train the convolution downsampling module and the LoRA adapter.
Since the bulk of the parameters ($\theta_{AE}$ and $\theta_{LLM}$) are never updated throughout the training process, it makes our framework compute efficient and allows us to scale to a large number of diverse datasets and tasks with limited compute resources. Moreover, it enables leveraging the existing capabilities of both pre-trained audio and language models without catastrophic forgetting. 
% Therefore, both $\theta_{AE}$ and $\theta_{LLM}$ are frozen throughout our training 
%
However, when training both
the downsampling module and the LoRA adapter from scratch
on a diverse set of speech tasks,
we observe frequent gradient explosion,
leading to suboptimal convergence.
Hence, we carefully design a curriculum of two stages for training.

In the first stage, we only train the convolution downsampling module
and the intermediate layer weights
without introducing the LoRA adapters.
Further, only the samples from the automatic speech recognition (ASR) task
are used in this stage. 
Since the encoded speech feature vectors can be very different from the token embeddings of the text input, 
this stage can help align them more easily in a common embedding space by only learning the 
parameters for the convolution downsampling module in the confined task space of ASR. 
This enables a pre-trained text-based LLM to attend to the content of the audio sequence 
and generate the speech transcription.

In the second stage, we now introduce the LoRA adapters
for training the model.
In this stage, 
the intermediate layer weights,
the downsampling module 
as well as the LoRA adapters
are unfrozen.
Since the LoRA adapters are training from scratch,
we first allow the adapter weights to warmup
by training only on the ASR task,
so as to get aligned to the common embedding space
learned by the convolution downsampling module in the first stage.
Finally, we introduce additional tasks on top of the ASR task
and continue training
while keeping the pre-trained audio encoder and LLM weights frozen.
Since the warmup using only the ASR task enables the model
to understand the contents of the audio,
our curriculum learning approach leads to
faster convergence on a variety of speech-based tasks
that rely on the spoken contents of the audio.

%% file: 400.experiments.tex
\section{Experiments}
\label{sec:experiments}

\subsection{Tasks}
\label{subsec:tasks}

\input{table-task-dataset-v2}

% general
In this work, we use a large collection of publicly available speech datasets from a diverse set of tasks. A summary of the datasets and evaluation metrics for these tasks is provided in the Table \ref{tab:multitask_datasets}, while examples and prompts are covered in the Table \ref{tab:training-task-qual}. Our training tasks include automatic speech recognition (ASR), five spoken language understanding (SLU) tasks, and five paralinguistic speech processing (PSP) tasks. The SLU tasks include those tasks which can be solved by a cascaded system of a ASR model and an LLM, while PSP tasks are classification tasks based on the audio, typically used in audio analytics.
% In the ST task, we train our models to translate audios with English speech to German, French and Romanian language text. 
For the IC/SL tasks, we split the SLURP dataset into seen and unseen intent/slot label classes and study them separately to understand the generalization capabilities of the model. The KWE task is about finding important keywords from the audio, while in the KWS task, we learn to classify whether a particular keyword was present in the audio or not. The target labels were synthetically created for both these tasks using an LLM. All other tasks are standard and an interested reader can refer to the Appendix \ref{sec:app-task} for more details.
% The PSP tasks include 5 classification tasks. Emotion Recognition (ER) is learning to classify the tone of the speaker into happy, sad, neutral or angry while Audio Sentiment Classification (ASC) classifies the speech as positive, negative or neutral. Speaker Counting (SC) task is to find if there were one or two speakers in an audio clip. In the Accent Classification (AC) task, we learn to classify the speaker's accent into five accent classes for English language. And finally, Speech/Non-Speech Detection (SNS) task is about identifying if speech was present in an audio segment. 
We create a list of at least 15 prompts per task describing the goal of the task. To further add diversity to the set of tasks, we use a text-to-speech (TTS) version of the Alpaca dataset \cite{taori2023stanford}.  This dataset contains a diverse collection of {prompt, input, output} tuples, where the prompt describes the task, input is the input for the task, and the output contains the target labels. However, there are no corresponding audios associated with the dataset. As in the existing work \cite{wang2023slm}, we use a TTS system (AWS Polly in our case) to generate synthetic audios for the input text using a pool of 10 different speakers. 
% This dataset adds to the diversity of the tasks and prompts used for training our models and enables generalization to new tasks. 

% The table also provides the details of evaluation metrics used for each of these tasks. For each task, we create a set of at least 15 prompts $\mathcal{P}^{\tau}$ by rephrasing the task descriptions. These prompts help the model learn to follow textual instructions, and enables generalization to new instructions, as discussed in Section \ref{subsec:generalization}. 
 % These audios serve as inputs $\mathbf{x}^\tau$, instructions as prompts $\mathcal{P}^{\tau}$, and outputs as targets $\mathbf{y}^\tau$.

\subsection{Models}
We train three different variants of multimodal models using our SpeechVerse framework, namely, 
(1) \textbf{Task-FT}: These represent a set of models where each model is trained individually for a particular task $\tau$. 
While most tasks have separate model, datasets from certain related tasks like IC/SF, KWE/KWS, and ER/ASC are trained together. 
(2) \textbf{Multitask-WLM}: This represents a single multi-task model trained by pooling datasets for all the tasks together. Both (1) and (2) uses a pretrained WavLM Large \cite{Chen2021WavLMLS} as the backbone audio encoder and only uses the last encoder layer output $AE(\mathbf{x}) = \mathbf{h}^{(L)}$ as the representation of the audio. 
(3) \textbf{Multitask-BRQ}: This model is similar to (2), but it uses the Best-RQ \cite{chiu2022self} architecture for the audio encoder. Because the Best-RQ encoder is trained using a random projection quantizer, the middle layer weights are more suited for downstream tasks if the encoder is frozen during fine-tuning. We therefore used a unified representation that combines representations from all layers via learnable weights for the multitask model trained with the BEST-RQ encoder. The details about the pretraining of this audio encoder are described in the Appendix \ref{sec:app-audio-encoder}. All the three variants of our models are trained using curriculum learning as highlighted in the Section \ref{sec:training}. All the models use Flan-T5-XL \cite{chung2022scaling} as the backbone LLM. The LoRA adapters introduced in Task-FT models have rank $(r) = 8$, while multitask models use $r=16$, allowing more learnable parameters for solving a diverse collection of tasks. The complete list of hyper-parameters and training setting is provided in the Appendix \ref{sec:app-hparam}.

\noindent\textbf{Baselines}:  For the SLU tasks, we compare our models with a cascaded baseline that uses an LLM on ASR hypotheses (ASR $\rightarrow$ LLM). For a fair comparison, we use a parameter-efficient fine-tuned version of Flan-T5-XL as the LLM for the baseline. The multi-task fine-tuning data is exactly the same between our models and the baseline except that the latter uses ground truth text in place of the audios. We benchmark the cascaded approach with ASR hypotheses from (1) a strong publicly available Whisper-large-v2 \cite{radford2023robust} ASR model and (2) our ASR Task-FT SpeechVerse model, enabling a true comparison between a multimodal model vs cascaded approach. 
Finally, we also benchmark the performance of the oracle ASR system by passing the ground truth transcripts to the baseline LLM (GT $\rightarrow$ LLM). For the KWS task, we use substring search of the keyword in ASR hypotheses as the baseline. For PSP tasks, we train task-specific classifiers that use the last layer representations from WavLM Large. The classifier contains a feed-forward layer, followed by a 2-layer Gated Recurrent Unit (GRU) with mean pooling over frames, followed by another 2 layers of feed-forward network and finally a softmax operator. These models are trained on the same task-specific data thereby allowing for direct comparison with our WavLM-based multimodal models.

\input{table-training-tasks-results}

%\input{table-qualitative}

%\vspace{-0.4cm}
% \subsection{Constrained Decoding}

% The work in \cite{willard2023efficient} introduced a model-agnostic technique to enforce domain-specific knowledge and constraints during text generation. Building on this prior approach, we have explored applying decoding constraints to the SpeechVerse model to improve generalization to unseen speech classification tasks. Rather than allowing the model to generate freely in response to a prompt, the decoding is restricted to output from a predefined vocabulary of class names. For example, in an intent classification task, the model would be constrained to only generate intent labels such as ``\textit{play\_radio}'', ``\textit{datetime\_query}'' or ``\textit{cooking\_recipe}''. By limiting the output space, the model is more likely to produce the desired class label rather than unrelated text.
%
% In this work, we also study the utility of this decoding strategy to improve generalization to unseen tasks.

\section{Results}

\subsection{Evaluation of SpeechVerse models}
\label{subsec:speechverse_performance}

We evaluate end-to-end trained joint speech and language models (E2E-SLM) leveraging the SpeechVerse framework on 11 unique tasks across multiple domains and datasets. We first evaluate SpeechVerse's core speech understanding capability through ASR benchmarks. We then evaluate more complex SLU tasks and paralinguistic speech tasks in Tables \ref{tab:pipelined_tasks} and \ref{tab:non_pipelined_tasks} respectively. %Finally, we compare our model performance against SOTA models on five representative tasks. 

\subsubsection{Performance on ASR and SLU tasks} 
\label{ssub:slu tasks}

First, we evaluate the performance of SpeechVerse models on four public ASR benchmark datasets namely \textit{libri-test-clean}, \textit{libri-test-other}, \textit{Voxpopuli} and \textit{CommonVoice}. The WER numbers for each of these datasets are reported in Table \ref{tab:pipelined_tasks}. 
SpeechVerse ASR in row 2 uses same model as task-specific pretrained ASR model (\textit{Task-FT}) in row 3.
When comparing our task-specific pretrained ASR model, which also serves as the initialization for multi-task finetuning, to Whisper ASR, our model achieves slightly better performance on average. However, the WER increases in both multitask models with \textit{Multitask-WLM} performing similarly to Whisper across three out of the four test sets. The lower performance of the multi-task SpeechVerse model compared to the task-specialized model is likely due to giving lower weight to ASR datasets when constructing batches during multi-task training. This was done to balance the performance across all tasks, since the data distribution is imbalanced between the different tasks.

When it comes to SLU tasks, a frequent question posed is if an end-to-end model can outperform a cascaded pipeline that transcribes speech via ASR and then feeds it to a language model. To investigate this, we conducted experiments on five semantic understanding tasks using the same foundation models as SpeechVerse. The text foundation model was further fine-tuned on data from the five SLU tasks separately, as we found the zero-shot performance of Flan-T5 on these benchmark test sets to be quite poor. We also report performance when feeding ground-truth transcripts into the fine-tuned LLM, to provide upper bound results. On 4 of the 5 tasks, excluding Keyword Extraction, the end-to-end trained models outperform the cascaded pipeline. In particular, the more commonly used tasks like intent classification, slot labeling, and speech translation are performing better than the cascaded system, demonstrating the efficacy of our models trained using SpeechVerse. We also observed that SpeechVerse models on KWS task are outperforming cascaded pipeline by an absolute 10\% in accuracy, while performing significantly behind on KWE task. Since the keyword search task requires an attention span focused on a specific word of interest, joint modeling helps improve accuracy by overcoming error propagation present in a cascaded pipeline. We also conducted an ablation study to determine if the KWE task benefits further from joint decoding of ASR transcription and keywords. We noticed an improvement in performance, closing the gap to the cascaded pipeline. The results from this study are detailed further in the sub-section \ref{ssub:compound_goals}. When comparing the multi-task models to the task-specific SpeechVerse models, there is a minor degradation in performance, but the difference is not substantial. Overall, the multitask model trained with either WavLM encoder or Best-RQ encoder outperformed cascaded systems in majority tasks.

\subsubsection{Performance on paralinguistic tasks} 
\label{subsec:plp_tasks}

The results in Table \ref{tab:non_pipelined_tasks} demonstrate clear improvements in performance on various paralinguistic speech processing tasks when using multi-task learning compared to fine-tuning the WavLM model independently for each task. Specifically, the SpeechVerse model trained with multitask learning using Best-RQ audio encoder (Multitask-BRQ) achieves gains over the baseline WavLM model of 4.8\% absolute on emotion recognition, 6.6\% on audio sentiment classification, and 2.5\% on accent classification. More modest gains are seen with the SpeechVerse model trained using multitask learning with WavLM encoder (Multitask-WLM). The unified representation's adaptive combination of all encoder layers helps multitask BEST-RQ model improve diverse paralinguistic task performance. Overall, multi-task learning provides noticeable improvements in model generalization and effectiveness across a diverse set of speech tasks compared to task-specific fine-tuning of the baseline WavLM model. The results highlight the advantages of learning shared representations across related tasks using multi-task learning techniques. 

% Additionally, we also benchmark the performance of our models against prior state-of-the-art models on five representative tasks - ASR, ST, IC, SL and ER - using public test sets. Overall, the SpeechVerse model demonstrated competitive performance compared to prior specialized models, the details of which are provided in Appendix \ref{app:sota_performance}. 

\subsubsection{Comparison against SOTA models}
\label{app:sota_performance}

\input{table-zero-shot-results}

Table \ref{tab:sota_benchmarking} benchmarks SpeechVerse models against state-of-the-art (SOTA) models on five diverse tasks: automatic speech recognition (ASR), speech translation (ST), intent classification (IC), slot filling (SF), and emotion recognition (ER). Across these tasks, SpeechVerse demonstrates competitive or superior performance compared to prior specialized models. When comparing our task-specific pretrained ASR model, which also serves as the initialization for multi-task finetuning, to Whisper ASR, our model achieves slightly better performance on average. However, the multitask model (\textit{Multitask-WLM}) performed similarly to Whisper across three out of the four test sets. When evaluating on speech translation across three language pairs, the task-specialized SpeechVerse model surpassed SeamlessM4T on two pairs, while the multi-task SpeechVerse model achieved competitive performance compared to prior work on average. Both models did not perform well on English to Romanian pair. The overall performance of the SpeechVerse models on speech translation is heavily limited by the capabilities of the underlying language model FlanT5. The speech translation capabilities cannot exceed the translation quality provided by FlanT5 as the base language model. To evaluate SpeechVerse on spoken language understanding tasks like intent classification (IC) and slot filling (SF), we retrained the task-specialized SpeechVerse model by incorporating all 69 intents (both seen and unseen) as well as all slots. This allowed us to compare SpeechVerse to prior work on the complete intent and slot sets. Our SpeechVerse model achieved competitive performance to the previous SOTA (\textit{PF-hbt-large}) on slot filling, but was significantly behind on intent classification with 5\% lower absolute accuracy. However, SpeechVerse outperformed the same SOTA model (\textit{Frozen-hbt-large}) by 10\% when the encoder weights were frozen during fine-tuning. To further analyze the gap to prior state-of-the-art, we conducted an experiment allowing the audio encoder weights to be tunable during fine-tuning. This achieved 89.5\% accuracy, matching the prior SOTA. This suggests the intent classification performance can overfit to the specific acoustic conditions of the SLURP dataset when full fine-tuning is performed. The SpeechVerse model that was trained end-to-end specifically for the task of emotion recognition achieved an 8\% absolute improvement in unweighted average recall over the previous state-of-the-art model (\textit{w2v2-L-robust}). In contrast, the multitask SpeechVerse model performed 3\% better than the prior state-of-the-art. However, one key difference is that the previous SOTA work trained on the MSP-Podcast 1.7 dataset, while we used version 1.11 for training. The test set version remained the same between the two approaches. Overall, the SpeechVerse model demonstrated competitive performance compared to prior specialized models in some cases when evaluated across the various tasks.

%Table \ref{tab:non_pipelined_tasks} reports performance on five paralinguistic classification tasks, comparing SpeechVerse to conventional baselines using a \textit{WavLM-large} speech encoder fine-tuned with a softmax classifier. Overall, SpeechVerse matches or exceeds the WavLM baseline across all tasks. Among the multitask models, combining representations from all layers significantly outperforms using just the top layer. Additionally, the multitask SpeechVerse models perform on par with or better than the task-specific SpeechVerse variants.

% \subsubsection{Comparison against SOTA models} Table \ref{tab:sota_benchmarking} benchmarks SpeechVerse against state-of-the-art models on five diverse tasks: automatic speech recognition (ASR), automatic speech translation (AST), intent classification (IC), slot filling (SF), and emotion recognition (ER). Across these tasks, SpeechVerse demonstrates competitive or superior performance compared to prior specialized models. 

\subsection{Generalization Across Instructions}
\label{subsec:generalization}
We comprehensively study our Multitask-WLM model's ability to generalize to diverse forms of unseen instructions. As a first, we try to accomplish seen tasks with differently worded instructions than those used for training. We create novel prompts for some of the training tasks and evaluate the robustness of the model to variations in the prompt. Next, we demonstrate the model's potential to leverage the robust language understanding of the underlying LLM to generalize to completely new tasks that the model has not seen at all during multimodal finetuning.
% We observe minimal variation in performance with the choice of prompts, indicating that the model can generalize well on its core capabilities. 

\subsubsection{Measuring robustness to prompt variations}
\label{app:prompts_performance}

\begin{table}[t]
\caption{Generalization to unseen prompts: The performance of each task is assessed on three different prompts, out of which, two are unseen during training.}
\label{tab:unseen_prompts}
\centering
\footnotesize
%\resizebox{0.49\textwidth}{!}{
\begin{tabular}{ccccc}
\toprule
\multirow{2}{*}{\textbf{Task}} & \multirow{2}{*}{\textbf{Dataset}} & \multirow{2}{*}{\makecell{\textbf{Prompt} \\ \textbf{(Unseen)}}}  & \multicolumn{2}{c}{\textbf{Performance}}  \\
\cmidrule{4-5}
& & & \textbf{Metric} & \textbf{Results} \\
\midrule
\multirow{3}{*}{ASR} & \multirow{3}{*}{Voxpopuli} & P1 (\xmark) & \multirow{3}{*}{WER$\downarrow$} & 6.8 \\
 & & P2 (\cmark) & & 7.1 \\
 & & P3 (\cmark) & & 6.8 \\
 \midrule
 \multirow{3}{*}{ST} & \multirow{3}{*}{\makecell{EuroParl \\ EN$\rightarrow$FR}} & P1 (\xmark) & \multirow{3}{*}{BLEU$\uparrow$} & 33.7 \\
 & & P2 (\cmark) & & 33.8 \\
 & & P3 (\cmark) & & 33.5 \\
 \midrule
 \multirow{3}{*}{AC} & \multirow{3}{*}{MCV} & P1 (\xmark) & \multirow{3}{*}{UAR$\uparrow$} & 60.0 \\
 & & P2 (\cmark) & & 60.7 \\
 & & P3 (\cmark) & & 60.6\\
\bottomrule
\end{tabular}
%}
% \vspace{-0.5cm}
\end{table}
To evaluate the effect of different prompts on the training task performance, we tested our MultiTask-WLM model with additional prompts for 3 different tasks: ASR, ST and AC. We test with three prompts per task, where one is directly taken from the set of prompts used during training, while two additional novel prompts are created by using a different wording and context. As showcased in the Table \ref{tab:unseen_prompts}, the model showecased similar performance across prompts for each task. For the ASR task, we only see a minor variation of 0.3 in WER between seen and unseen prompts. Similarly for the ST and the AC task, the metric values deviates very little with the change in prompt. These small differences indicate that our model has generalized its core capabilities well and is not strongly dependent on the specific wording or context of the prompts. While prompt engineering remains important for optimal performance, these results suggest that the model has learned robust task-specific skills that transfer across the choice of prompts, at least for the tasks seen during training. For the tasks beyond those seen during training, we study the prompt design separately in a later subsection \ref{ssub:constrained_decoding}.

\subsubsection{Open-ended Instruction Following}
\input{table-qualitative}

\label{ssec:open_ended}

To study the model's ability to understand open-ended text-based as well as speech-based instructions, we prompted the model with several unrestricted creative requests that were not a part of our training curriculum.
We enumerate several such examples in the Table \ref{tab:qualitative_examples}. In many of these examples, the model is required to exhibit
profound comprehension of both the spoken and written directives to successfully execute the task.
For example, in the Creative QA task, the model has to understand the spoken request as well as the instruction prompt
in order to generate a related response.
In the Summarization task, the model has to correctly surmise the spoken content to generate a summary.
In the Contextual Biasing task, we observe that the model is even able to correct its own output when provided with hints.
The robust responses of the multi-task model with such a distributional shift in the input from the training data 
demonstrates the adaptability of its core instruction following skills. Rather than overfitting to the training domain, the multi-task learning approach  enables the model to learn more versatile capabilities  in instruction comprehension and execution that better transfer to new contexts. We provide some quantitative results on unseen tasks and labels in the next section. 

\subsection{Strategies for Improving Performance}
We further evaluate strategies to improve the multi-task model's performance specially for unseen tasks and class labels. First, we leverage contrained decoding \cite{willard2023efficient} for tasks that have a pre-defined set of finite outcomes. Next, we also study joint decoding of the output of the task with the ASR hypotheses of the audio for certain complex spoken language understanding tasks. 

\input{table-generalization}
\subsubsection{Constrained Decoding}
\label{ssub:constrained_decoding}
The work in \cite{willard2023efficient} introduced a model-agnostic technique to enforce domain-specific knowledge and constraints during text generation. Building on this prior approach, we have explored applying decoding constraints to the SpeechVerse model to improve generalization to unseen speech classification tasks. Rather than allowing the model to generate freely in response to a prompt, the decoding is restricted to output from a predefined vocabulary of class names. For example, in an intent classification task, the model would be constrained to only generate intent labels such as ``\textit{play\_radio}'', ``\textit{datetime\_query}'' or ``\textit{cooking\_recipe}''. By limiting the output space, the model is more likely to produce the desired class label rather than unrelated text.

We meticulously benchmark the model's performance on
close-ended tasks, such as a diverse set of classification tasks,
that have a pre-defined set of finite class labels.
To understand the influence of the instruction prompts,
we divide this study into two parts:
(1) where we only provide the class labels in the prompt,
and (2) where we provide an accompanying description of each class label in the prompt.
We ensure that none of these class labels were seen during training,
and hence these are all novel tasks for the model.
Further, we evaluate the efficacy of employing constrained decoding
in each of these two parts,
as the class labels are known to us beforehand.
Note here that the SL task can be considered a harder task
as the model has to correctly classify a slot label
as well as identify the corresponding slot value from the speech.
Hence, we report both, the SLU-F1 metric as well as the SD-F1 (Slot label Detection) metric for SL.
The results of this study are presented in Table \ref{tab:constrained_decoding}.

We observe that including descriptions in the prompt has inconsistent results,
which can be attributed to the quality and subjectivity of the descriptions provided in the prompt,
especially as these descriptions were not seen during training.
However, we see that constrained decoding improves upon the results in all cases,
and most significant gains are observed only when descriptions are provided
with constrained decoding.
This indicates that providing descriptions 
indeed steers the model towards 
better comprehension of the task semantics,
but only constrained decoding is able
to objectively prune the noise introduced
by any prompt bias.
This phenomena is further revealed in the SL task,
where the SLU-F1 has a lower absolute value compared to SD-F1, as
the SLU-F1 metric incorporates both slot label and slot value,
whereas constrained decoding can only be applied to the slot label 
(hence the higher SD-F1). Similarly, for a completely unseen task of Domain Classification (DC), where the goal is to classify the content of the audio into five domains like healthcare, technology etc, we observe a strong performance of 62\% accuracy with constrained decoding.

\subsubsection{Joint Decoding}
\label{ssub:compound_goals}
Certain SLU tasks require the model to understand the semantics of the audio or perform a operation on the content of the audio. For example, KWE task is about extracting important keywords from the ASR hypothesis of the audio. 
% As highlighted in the Section \ref{ssub:slu tasks}, we see a worse performance for KWE task with our multimodal models as compared to cascaded baseline. This task requires the model to understand the semantics of the audio, identify the useful words and produce them as keywords in the output. 
Since this is a multi-step reasoning process for the model, we take inspiration from the existing work \cite{wei2022chain} on Chain-of-Thought (CoT) prompting. We train our model to first decode the ASR hypothesis of the audio, followed by the output of the task. The prompts used for the joint elicitation of ASR hypothesis and the task output are described in the Table \ref{tab:compound_goal_example}. For a representative set of SLU tasks including IC, KWE and ER, we re-train the Task-FT models by adding a small portion of such multi-step examples along with single-task examples. We compare the results with and without joint decoding with ASR hypothesis in the Table \ref{tab:compound_goals}.

\begin{table*}[ht]
\caption{Example prompt for elicitation of compound goals for KWE and ER task}
\label{tab:compound_goal_example}
% \vspace{-2mm}
\centering
\footnotesize
%\resizebox{0.49\textwidth}{!}{
\begin{tabular}{L{0.04\textwidth} L{0.59\textwidth} L{0.28\textwidth}}
\toprule
\textbf{Task} & \textbf{Instruction} & \textbf{Prediction}\\
\midrule
\multirow{1}{0.05\textwidth}{KWE} & Perform the following audio-based tasks in the order as described. \newline === Task: ASR === \newline Perform speech recognition using the preceding audio. \newline === Task: KWE ===\newline Identify significant keywords in the provided audio.\newline Make sure to format the output as "ASR: ... | KWE: ... |" &  ASR: paris is the capital of france | KWE: paris, france | \\
\midrule
\multirow{1}{0.05\textwidth}{ER} & Perform the following audio-based tasks in the order as described.\newline === Task: ASR ===\newline What is being said in the audio? \newline === Task: Emotion ===\newline Classify the tone of the speaker as happy, sad, angry or neutral \newline Make sure to format the output as "ASR: ... | Emotion: ... |" & ASR: can you shut up for a while | Emotion: angry |\\
\bottomrule
\end{tabular}
%}
\end{table*}

\newcolumntype{L}[1]{>{\raggedright\let\newline\\\arraybackslash\hspace{0pt}}m{#1}}

\begin{table}[h]
\caption{Results on compound goal experiments when decoding \\ performed with (w/) and without (w/o) ASR.}
\label{tab:compound_goals}
% \vspace{-2mm}
\centering
\footnotesize
%\resizebox{0.49\textwidth}{!}{
\begin{tabular}{ccccc}
\toprule
\multirow{3}{*}{\textbf{Task}} & \multirow{3}{*}{\textbf{Dataset}} &  \multirow{3}{*}{\textbf{Metric}} & \multicolumn{2}{c}{\makecell{\textbf{Joint Decoding}}} \\
\cmidrule{4-5}
& &  & \textbf{w/o} & \textbf{w/} \\
\midrule
\multirow{1}{*}{IC} & SLURP  & \multirow{1}{*}{ACC$\uparrow$} & 90.4 & 92.2 \\
\midrule
\multirow{2}{*}{KWE} & MCV  & \multirow{2}{*}{F1$\uparrow$} & 46.9 & 49.5 \\
 & Vox  &  & 53.1 & 56.7 \\
\midrule
ER & MSP  & UAR$\uparrow$ & 64.5 & 64.7 \\
\bottomrule
\end{tabular}
%}
\end{table}

The results in the table showcase that augmenting the training data with compound goals helps to improve the performance for all three tasks. The improved performance can be attributed to the possibility of self-attention on the already decoded ASR hypothesis in the decoder of our multimodal model. Further, such multi-step training examples brings out the true multimodal capabilities of the model to successfully complete the combination of tasks. Further, such a paradigm can help save crucial inference latency by using a single call of the large multi-modal model for obtaining both the transcript and the task output. A more detailed analysis on understanding the benefits of joint decoding will be conducted in future work. 
% This improved feature sharing then transfers to the individual tasks, demonstrating how compound goals can enhance multi-task learning. Further research is needed to fully understand the interaction effects between the different data types.

% For instance, a model trained on speech recognition, intent classification, and compound examples integrating both tasks, achieves higher F1 on the intent classification task compared to training without compound examples. This indicates the model is learning a more generalized representation of the speech input that transfers better to the individual task. The compound examples force the model to simultaneously learn the multiple skills, acting as a form of hard task augmentation. %\pra{OR self-attention in decoder affects performance on both tasks? Make this more specific.}

% Overall, supplementing multi-task data with limited compound goal training signals appears to provide a beneficial regularization effect.

\input{training-tasks-qualitative}

%% file: table-task-dataset-v2.tex
\begin{table*}[t]
\caption{Details of our tasks, training datasets and evaluation metrics. ST, IC, SF, KWE and KWS tasks are referred as spoken language understanding (SLU) tasks, while ER, ASC, SC, AC and SNS represents paralinguistic speech processing (PSP) tasks}
\label{tab:multitask_datasets}
\centering
\footnotesize
\resizebox{0.99\textwidth}{!}{
\begin{tabular}{ccccrr}
\toprule
\multirow{2}{*}{\textbf{Task}} & \multirow{2}{*}{\textbf{Description}} & \multirow{2}{*}{\textbf{Metric}} & \multirow{2}{*}{\textbf{Dataset}} & \multicolumn{2}{c}{\textbf{Training}}   \\
& & & & \textbf{\#Hours} & \textbf{\#Samples} \\
\midrule
\multirow{5}{*}{ASR} & \multirowcell{5}{Automatic Speech \\ Recognition} & \multirowcell{5}{Word Error \\ Rate (WER)$\downarrow$} & Librispeech  \cite{panayotov2015librispeech} & 960 & 281K \\
 & & & Mozilla Common Voice 5.1 \cite{ardila2019common} & 1.4K & 1.1M \\
 &  & & VoxPopuli \cite{dekel2009vox} & 385 & 160K  \\
  &  & & SLURP \cite{bastianelli2020slurp} & 84 & 119K \\
  % &  & & SLURP \cite{bastianelli2020slurp} & 84 & 119K & 10.2 & 1.3K \\
   % &  & & EuroParl \cite{koehn2005europarl} & 92 & 39K& 3 & 1.3K \\
   &  & & EuroParl \cite{koehn2005europarl} & 92 & 39K\\
    &  & & MSP-Podcast 1.11 \cite{lotfian2017msp} & 135 & 83K \\
 \midrule
 \multirow{2}{*}{ST} & \multirow{2}{*}{Speech Translation} & \multirow{2}{*}{BLEU $\uparrow$} & CoVost2 \cite{wang2020covost} & 426 & 576K  \\
 & & & EuroParl \cite{koehn2005europarl} & 73 & 90K  \\
 \midrule
 IC & Intent Classification & Accuracy (ACC)$\uparrow$ & SLURP \cite{bastianelli2020slurp}& 35 & 47K \\
 SF & Slot Filling &  SLU-F1 $\uparrow$ & SLURP \cite{bastianelli2020slurp} & 25 & 30K \\
 \midrule
 \multirow{2}{*}{KWE} & \multirow{2}{*}{Keyword Extraction} & \multirow{2}{*}{Macro F1$\uparrow$} & Librispeech \cite{panayotov2015librispeech} & 960 & 281K \\
 & &  & Mozilla Common Voice 5.1 \cite{ardila2019common} & 426 & 288K \\
 \midrule
 \multirow{2}{*}{KWS} & \multirow{2}{*}{Keyword Search} & \multirowcell{2}{Accuracy (ACC) $\uparrow$} & Librispeech \cite{panayotov2015librispeech} & 960 & 281K \\
 & & & Mozilla Common Voice 5.1 \cite{ardila2019common} & 426  & 288K \\
 \midrule
 ER & Emotion Recognition & \multirowcell{5}{Unweighted Average\\ Recall (UAR) $\uparrow$} & MSP-Podcast 1.11 \cite{lotfian2017msp}& 91 & 57K \\
 ASC & Audio Sentiment Classification & & MSP-Podcast 1.11 \cite{lotfian2017msp}& 135 & 84K \\
 SC & Speaker Counting & & Fisher \cite{Fisher1, Fisher2} & 690 & 755K \\
 AC & Accent Classification & & Mozilla Common Voice 5.1 \cite{ardila2019common}& 190 & 123K \\
 SNS & Speech/Non-Speech Detection & & In-house VAD & 269 & 149K \\
\bottomrule
\end{tabular}
}
% \vspace{-0.5cm}
\end{table*}

%% file: table-training-tasks-results.tex
\begin{table*}[t]
\caption{Results of ASR and spoken language understanding 
(SLU) tasks. Datasets are defined as: LTC: Librispeech test-clean; LTO: Librispeech test-other; Vox: Voxpopuli; MCV: Mozilla Common Voice; EN: English; DE: German; FR: French;}
% \vspace{-0.2cm}
\label{tab:pipelined_tasks}
\centering
\footnotesize
\resizebox{\linewidth}{!}{
\begin{tabular}{cccccccccccc}
\toprule
% &  & \multicolumn{10}{c}{\textbf{Task (Metric)}} \\
%\cmidrule{3-10}
\multirow{4}{*}{\textbf{Model}} & \multirow{4}{*}{\textbf{Training}} & \multicolumn{4}{c}{\textbf{\makecell{ASR \\ (WER$\downarrow$)}}} & \textbf{\makecell{IC \\ (ACC$\uparrow$)}} & \textbf{\makecell{SF \\ (SLU-F1$\uparrow$)}} & \multicolumn{2}{c}{\textbf{\makecell{ST \\ (BLEU$\uparrow$)}}} & \textbf{\makecell{KWE \\ (F1$\uparrow$)}} & \textbf{\makecell{KWS \\ (ACC$\uparrow$)}} \\
\cmidrule{3-12}
\multicolumn{2}{c}{} & \textbf{LTC} & \textbf{LTO} & \textbf{Vox} & \textbf{MCV} & \textbf{SLURP} & \textbf{SLURP} & \textbf{EN$\rightarrow$DE} & \textbf{EN$\rightarrow$FR} & \textbf{MCV} & \textbf{MCV} \\
\midrule
% production numbers
% ASR $\rightarrow$ LLM & \multirow{2}{*}{\makecell{LLM-FT}} & & &  &   & 85.9 & 59.6 & - & 24.2 & 57.3 & 90.3 \\
% whisper numbers
Whisper ASR $\rightarrow$ LLM & \multirow{2}{*}{\makecell{LLM-FT}} & 2.5 & 4.9 & 7.0 & 8.2 & 86.6 & 72.3 & 23.5 & 30.6 & 45.1 & 87.7 \\
% audioLLM ASR numbers
SpeechVerse ASR $\rightarrow$ LLM & & 2.1 & 4.4 & 6.5 & 10.5 &   88.8 &  80.5 & 24.4 & 30.9 & 54.3 & 87.3 \\
\midrule
\multirow{3}{*}{\makecell{SpeechVerse \\ (E2E--SLM)}} & Task-FT &  2.1 & 4.4 & 6.5 & 10.5 &  90.4 & 82.4 & 27.8 & 35.7 & 46.9 & 98.7 \\
 & Multitask-WLM & 2.5 & 4.7 & 6.8 & 12.0 & 90.3 & 82.2 & 25.9 & 33.7 & 46.6 & 98.5 \\
 & Multitask-BRQ &  3.0 & 6.7 & 7.1 & 11.8 & 90.0  & 83.4 & 25.2 & 32.5 & 44.2 & 98.4 \\
  % & Multitask-All-cont &  2.9 & 6.2 & 7.0 & 11.6 & 89.8  & 83.9 & 25.6 & 32.8 & 45.0 & 98.7 \\
   % & Multitask-All-joint &   &  &  &  & 89.6  & 83.9 & 17 & 11 & 46.2 &  \\
 \midrule
GT $\rightarrow$ LLM & LLM-FT & - & - & - & - & 94.2 & 84.3 & 27.2 & 34.8 & 70.3 & 100 \\ 
\bottomrule
\end{tabular}
}
% \vspace{-0.2cm}
\end{table*}

\begin{table}[t]
\caption{Results of paralinguistic speech processing (PSP) tasks. All reported numbers are the value of the UAR metric.}
% \vspace{-0.2cm}
\label{tab:non_pipelined_tasks}
\centering
\footnotesize
\resizebox{0.6\linewidth}{!}{
\begin{tabular}{ccccccc}
\toprule
%& & \multicolumn{5}{c}{\textbf{Task (Metric)}} \\
%\cmidrule{3-7}
\multirow{3}{*}{\textbf{Model}} & \multirow{3}{*}{\textbf{Training}} & \textbf{ER} & \textbf{ASC} & \textbf{SC} & \textbf{AC} & \textbf{SNS} \\
\cmidrule{3-7}
& & \textbf{MSP} & \textbf{MSP} & \textbf{Fisher} & \textbf{MCV} & \textbf{VAD} \\
\midrule
WavLM & Task-FT & 60.3 & 57.5 & 98.5 & 57.9 & 96.3 \\
\midrule
\multirow{3}{*}{\makecell{SpeechVerse \\ (E2E--SLM)}}  & Task-FT & 64.6 &  61.3 & 99.6 & 60.1 & 96.9 \\
 & Multitask-WLM & 62.0 & 60.1 & 98.9 & 60.0 & 97.6 \\
 & Multitask-BRQ & 65.1 & 64.1 & 99.3 & 60.4 & 97.5 \\
  % & Multitask-All-cont &  &  & 99.4 & 59.1 & 98.8 \\
   % & Multitask-All-cont-joint & 64.5 & 60.2 &  &  &  \\
\bottomrule
\end{tabular}
}
% \vspace{-0.5cm}
\end{table}

%% file: table-zero-shot-results.tex
\begin{table}
\caption{Comparison of SpeechVerse models to prior specialized SOTA models on five diverse tasks: automatic speech recognition (ASR), speech translation (ST), intent classification (IC), slot filling (SF), and emotion recognition (ER).}
\label{tab:sota_benchmarking}
\centering
\footnotesize
\resizebox{0.9\linewidth}{!}{
\begin{tabular}{ccccc}
\toprule
\multirow{2}{*}{\textbf{Task}} & \multirow{2}{*}{\textbf{Dataset}} & \multirow{2}{*}{\textbf{Model}} & \multicolumn{2}{c}{\textbf{Performance}}  \\
\cmidrule{4-5}
 & & & \textbf{Metrics} & \textbf{Results} \\
\midrule
\multirow{12}{*}{ASR} & \multirow{4}{*}{\makecell{Librispeech \\ \textit{test-clean} | \textit{test-other}}} & Whisper large-v2 \cite{radford2023robust} & \multirow{3}{*}{WER $\downarrow$} & \makecell{2.5 | 4.9} \\
 & & SLM-FT \cite{wang2023slm} & & \makecell{2.6 | 5.0} \\
 % &  & Qwen-Audio \cite{chu2023qwen} &  & \makecell{2.0 | 4.2} \\
 &  & SpeechVerse Task-FT &  & \makecell{\textbf{2.1} | \textbf{4.4}} \\
  &  & SpeechVerse Multitask-WLM &  & \makecell{2.5 | 4.7} \\
\cmidrule{2-5}
 & \multirow{3}{*}{VoxPopuli} & Whisper large-v2 \cite{radford2023robust} & \multirow{3}{*}{WER $\downarrow$} & 7.0 \\
 & & mSLAM-CTC \cite{bapna2022mslam} & & 7.0 \\
  &  & SpeechVerse Task-FT &  &  \textbf{6.5}\\
  &  & SpeechVerse Multitask-WLM &  &  6.8\\
\cmidrule{2-5}
 & \multirow{3}{*}{CommonVoice 5.1} & Whisper large-v2 \cite{radford2023robust} & \multirow{3}{*}{WER $\downarrow$} & 8.2 \\
 & & SLM-FT \cite{wang2023slm} & & \textbf{7.5} \\
 &  & SpeechVerse Task-FT &  &  10.5\\
  &  & SpeechVerse Multitask-WLM &  &  12.0\\
\midrule
\multirow{4}{*}{ST} & \multirow{4}{*}{\makecell{EuroParl \\ EN$\rightarrow$DE | EN$\rightarrow$FR | EN$\rightarrow$RO}} & SeamlessM4T 1.2B~\cite{communication2023seamlessm4t} & \multirow{3}{*}{BLEU $\uparrow$} & 27.8 | 30.3 | \textbf{38.7} \\
 & & XMEF~\cite{li-etal-2021-multilingual} & & 22.5 | 30.0 | 32.3 \\ 
 & & SpeechVerse Task-FT & & \textbf{27.8} | \textbf{35.7} | 32.2 \\ 
  & & SpeechVerse Multitask-WLM & & 25.9 | 33.7 | 30.1 \\ 
\midrule
\multirow{4}{*}{IC} & \multirow{4}{*}{SLURP} & E2E-SLU CTI \cite{seo2022integration} & \multirow{4}{*}{ACC $\uparrow$} & 86.9 \\
& & Frozen-hbt-large \cite{wang2022fine} & & 74.4 \\
 & & PF-hbt-large \cite{wang2022fine} & & \textbf{89.2} \\ 
 & & SpeechVerse Task-FT$^*$ & & 84.6 \\ 
\midrule
\multirow{4}{*}{SF} & \multirow{4}{*}{SLURP} & E2E-SLU CTI \cite{seo2022integration} & \multirow{4}{*}{SLU-F1 $\uparrow$} & 74.7 \\
& & Frozen-hbt-large \cite{wang2022fine} & & 60.1 \\
 & & PF-hbt-large \cite{wang2022fine} & & \textbf{78.9} \\ 
 & & SpeechVerse Task-FT$^*$ & & 76.7 \\ 
 \midrule
 \multirow{3}{*}{ER} & \multirow{3}{*}{MSP-Podcast 1.7} & w2v2-L-robust\cite{derington2023testing} & \multirow{3}{*}{UAR $\uparrow$} & 58 \\
 & & SpeechVerse Task-FT & & \textbf{66.7} \\ 
  & & SpeechVerse Multitask-WLM & & 61.2 \\ 
\bottomrule
\multicolumn{5}{r}{\tiny $^*$SpeechVerse Task-FT model was re-trained for SLURP by including \textit{all} intents and slots for comparison with other models.}
\end{tabular}
}
\end{table}

% Prashant: Are these correct numbers?  
%  \cmidrule{2-5}
% \multirow{3}{*}{ST} & \multirow{3}{*}{\makecell{CoVoST \\ \textit{en-de} | \textit{en-ca}}} & Whisper large-v2~\cite{radford2023robust} & \multirow{3}{*}{BLEU $\uparrow$} &  \\
%  & & Qwen-Audio \cite{chu2023qwen} & &  \\ 
%  & & SpeechVerse & & \\ 

%% file: table-qualitative.tex
\newcolumntype{L}[1]{>{\raggedright\let\newline\\\arraybackslash\hspace{0pt}}m{#1}}

\begin{table}[]
\caption{Qualitative examples on open-ended tasks showcasing instruction-following ability of the model. DC: Domain Classification; QA: Question Answering; ST: Speech Translation}
\label{tab:qualitative_examples}
\centering
\footnotesize
%\resizebox{0.49\textwidth}{!}{
\resizebox{\linewidth}{!}{
\bgroup
\def\arraystretch{1.5}
\begin{tabular}{L{0.09\textwidth} L{0.46\textwidth} L{0.46\textwidth} }
\toprule
\textbf{Task} & \textbf{Instruction} & \textbf{Prediction} \\
\midrule
& \multicolumn{2}{l}{\includegraphics[width=15pt,valign=c]{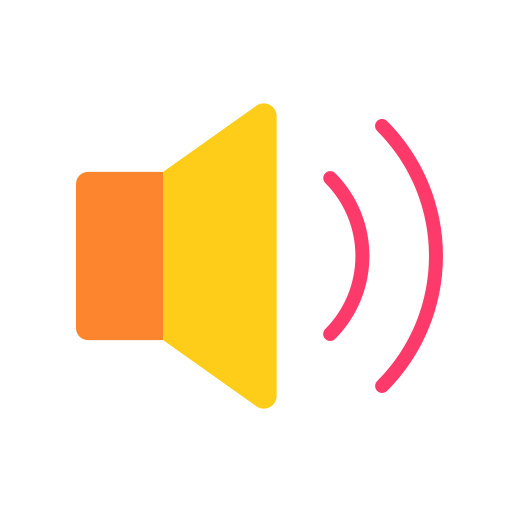} [Transcription: are you able to update the personal details on my account]} \\ \cmidrule{2-3}
\rowcolor{gray!15}
\multirow{2}{0.08\textwidth}{\cellcolor{white}Post Processing} & 
Transcribe the preceding audio & 
{\textit{are you able to update the personal details on my account}} \\
% \cdashline{2-3}
& Transcribe the preceding audio with punctuations & 
{\textit{are you able to update the personal details on my account?}} \\ \cmidrule{2-3}
\rowcolor{gray!15}
{\cellcolor{white}Response Generation} & Generate a response to the user query in the preceding audio & {\textit{Sure, I can update the personal details on your account. Can you please tell me which account you are referring to and the specific details you would like me to update?}} \\ \cmidrule{2-3}
{\cellcolor{white}ST \ \ \ \ \ \ {\tiny unseen language}} & Translate the preceding audio from english into spanish & {\textit{Puedes actualizar los datos personales de mi cuenta?}} \\
\midrule
& \multicolumn{2}{l}{\includegraphics[width=15pt,valign=c]{figures/audio-icon.png} [Transcription: i need synthroid supply that will last me a month]} \\ \cmidrule{2-3}
\rowcolor{gray!15}
\multirow{6}{0.08\textwidth}{\cellcolor{white}Contextual Biasing} & What is being said in the preceding audio? & {\textit{i need \textcolor[rgb]{0.7, 0.1, 0.1}{\textbf{cynthroid}} supply that will last me a month}}\\
& What is being said in the preceding audio? Hints: thyroxine, synthroid, thyroid, hormones & {\textit{i need synthroid supply that will last me a month}}\\
& {\cellcolor{gray!15}You are an AI assistant for a pharmaceutical company that specializes in thyroid treatment. Based on this context, what is being said in the preceding audio?} & {\cellcolor{gray!15}{\textit{i need synthroid supply that will last me a month}}}\\ \cmidrule{2-3}
DC & 
Classify the domain of the user audio input into finance, healthcare, travel, internet or food & 
{\textit{healthcare}} \\ \cmidrule{2-3}
\rowcolor{gray!15}
{\cellcolor{white}QA} & 
What is the name of the medicine in the preceding audio? & 
{\textit{Synthroid}} \\ 
% \cmidrule{2-3}
% Emotion & 
% Is the tone of the user neutral or angry in the audio? & 
% {\textit{neutral}} \\
\midrule
& \multicolumn{2}{l}{\includegraphics[width=15pt,valign=c]{figures/audio-icon.png} A 30-second audio generated using a text-to-speech system with contents of a news article$^*$} \\ \cmidrule{2-3}
\rowcolor{gray!15}
\multirow{3}{0.08\textwidth}{\cellcolor{white}Summar-\\ization} & Generate an abstractive summary for the content of the preceding audio in a single line &  {\textit{Special counsel jack smith urges us supreme court to let former president donald trump’s election interference case proceed to trial without further delay.}}\\
&  Create a very crisp newspaper headline for the preceding audio &  {\textit{Former President Donald Trump Faces Election Interference Charges in Court}}\\
\midrule

& \multicolumn{2}{l}{\includegraphics[width=15pt,valign=c]{figures/audio-icon.png} [Transcription: Can you explain the process of composting?]} \\ \cmidrule{2-3}
\rowcolor{gray!15}
{\cellcolor{white}Factual QA} & Generate a short factual answer to the question in the preceding audio &  {\textit{The process of composting is a gradual decomposition of organic matter into soil, with the goal being to decompose the organic matter into nutrients and water.}} \\
\midrule
& \multicolumn{2}{l}{\includegraphics[width=15pt,valign=c]{figures/audio-icon.png} [Transcription: Compose a descriptive paragraph about a childhood memory.]} \\ \cmidrule{2-3}
\rowcolor{gray!15}
{\cellcolor{white}Creative QA} & Generate a short creative answer to the question in the preceding audio &  {\textit{A child's imagination is a powerful tool for creating memorable images, and one of my favorite memories is the creation of a dollhouse complete with all the pieces.}} \\
\bottomrule
\multicolumn{3}{r}{\tiny $^*$\url{https://apnews.com/article/trump-special-counsel-election-interference-january-6-c2dcc83e56a541804d4785f6bb6cd45c}}
\end{tabular}
\egroup
}
% \vspace{-0.5cm}
\end{table}

%% file: table-generalization.tex
\begin{table}[t]
\caption{Results on three unseen tasks with and without constrained decoding (CD). We compare the settings wherein the instruction prompt only contained the class labels v/s when descriptions were provided for each class label in the prompt.}
\label{tab:constrained_decoding}
% \vspace{-2mm}
\centering
\footnotesize
\resizebox{0.8\linewidth}{!}{
% \begin{tabular}{@{}cclcclcclc@{}}
% \toprule
% \multirow{3}{*}{\begin{tabular}[c]{@{}c@{}}\\Class Label\\ Description\end{tabular}} & \multirow{3}{*}{\begin{tabular}[c]{@{}c@{}}\\CD\end{tabular}} &  & \multicolumn{2}{c}{IC} &  & \multicolumn{2}{c}{SL} &  & DC \\ \cmidrule(lr){4-5} \cmidrule(lr){7-8} \cmidrule(l){10-10} 
%  &  &  & SLURP$^*$ & SNIPS &  & \multicolumn{2}{c}{SLURP$^*$} &  & Internal \\
%  &  &  & (ACC $\uparrow$) & (ACC $\uparrow$) &  & (SD-F1 $\uparrow$) & (SLU-F1 $\uparrow$) &  & (ACC $\uparrow$) \\ \midrule
% \multirow{2}{*}{No} & No &  & 51.7 & 54.9 &  & 73.12 & 46.45 &  & 56.9 \\
%  & Yes &  & 68.8 & 68.9 &  & 73.38 & 46.49 &  & 59.0 \\ \midrule
% \multirow{2}{*}{Yes} & No &  & 47.7 & 61.0 &  & 69.79 & 47.34 &  & 44.0 \\
%  & Yes &  & \textbf{70.2} & \textbf{75.9} &  & \textbf{77.47} & \textbf{48.45} &  & \textbf{62.0} \\ \bottomrule
% \multicolumn{10}{r}{\footnotesize $^*$This is a subset of SLURP labels not seen during training}
% \end{tabular}
\begin{tabular}{@{}cccclcclc@{}}
\toprule
\multirow{3}{*}{\begin{tabular}[c]{@{}c@{}}Class Label\\ Description\end{tabular}} & \multirow{3}{*}{\begin{tabular}[c]{@{}c@{}}CD\end{tabular}} & \multicolumn{2}{c}{IC} &  & \multicolumn{2}{c}{SL} &  & DC \\ \cmidrule(lr){3-4} \cmidrule(lr){6-7} \cmidrule(l){9-9} 
 &  & SLURP$^*$ & SNIPS &  & \multicolumn{2}{c}{SLURP$^*$} &  & Internal \\
 &  & (ACC) & (ACC) &  & (SD-F1) & (SLU-F1) &  & (ACC) \\ \midrule
\multirow{2}{*}{No} & No & 51.7 & 54.9 &  & 73.12 & 46.45 &  & 56.9 \\
 & Yes & 68.8 & 68.9 &  & 73.38 & 46.49 &  & 59.0 \\ \midrule
\multirow{2}{*}{Yes} & No & 47.7 & 61.0 &  & 69.79 & 47.34 &  & 44.0 \\
 & Yes & \textbf{70.2} & \textbf{75.9} &  & \textbf{77.47} & \textbf{48.45} &  & \textbf{62.0} \\ \bottomrule
\multicolumn{9}{r}{\scriptsize $^*$This is a subset of SLURP labels not seen during training}
\end{tabular}
}
\end{table}

%% file: training-tasks-qualitative.tex
% \newcolumntype{L}[1]{>{\raggedright\let\newline\\\arraybackslash\hspace{0pt}}m{#1}}

\begin{table*}[ht]
\caption{Qualitative examples of output from all the training tasks on a audio file with transcription "turn on the radio now"}
\label{tab:training-task-qual}
\centering
% \footnotesize
%\resizebox{0.49\textwidth}{!}{
\resizebox{\linewidth}{!}{
\bgroup
\def\arraystretch{1.5}
\begin{tabular}{L{0.09\textwidth} L{0.70\textwidth} L{0.36\textwidth} }
\toprule
\textbf{Task} & \textbf{Instruction} & \textbf{Prediction} \\
\midrule
\rowcolor{gray!15}
\multirow{2}{0.08\textwidth}{\cellcolor{white}ASR} & 
Transcribe the preceding audio & 
{\textit{turn on the radio now}} \\
& What is being said in the preceding audio? & 
{\textit{turn on the radio now}} \\ \midrule
\rowcolor{gray!15}
\multirow{2}{0.08\textwidth}{{\cellcolor{white}ST}} & Translate the english audio to german & {\textit{schalten sie jetzt den radio auf}} \\ 
& Generate french translation for the english audio & {\textit{mettez le radio à l'onder maintenant}} \\ 
\midrule
\rowcolor{gray!15}
\multirow{1}{0.08\textwidth}{{\cellcolor{white}IC}} & The previous audio needs to be mapped to exactly one of the following intents: \newline
- datetime\_query: user asks about date, time, schedules, etc; \newline
- qa\_definition: user wants to ask for general fact-based questions to receive definitions, explanations, descriptions, etc; \newline
- calendar\_remove: user wants to remove calendar events by voice; \newline
... \newline
% ... \newline
- play\_radio: user wants to request playback of radio, stations, etc; 
& {\textit{play\_radio}} \\ 
\midrule
\multirow{1}{0.08\textwidth}{{\cellcolor{white}SF}} & Identify any of the following slots present in the speech: \newline - date: any reference to a date, month or day of the week; \newline - place\_name: any reference of a place e.g., berlin, new york etc.; \newline - person: any reference of a person; \newline - time: any reference to time e.g., three am, two thirty etc.;
& {\textit{time = now;}} \\ 
\midrule
\rowcolor{gray!15}
\multirow{1}{0.08\textwidth}{{\cellcolor{white}KWE}} & Identify important keywords in the preceding recording & {\textit{radio}} \\ 
\midrule
\multirow{1}{0.08\textwidth}{{\cellcolor{white}KWS}} & Is the word 'radio' used in the audio? & {\textit{yes}} \\ 
\midrule
\rowcolor{gray!15}
\multirow{1}{0.08\textwidth}{{\cellcolor{white}ER}} & Detect the primary emotion conveyed in this audio by the speaker's tone - is it happiness, sadness, anger, or neutral? & {\textit{neutral}} \\ 
\midrule
\multirow{1}{0.08\textwidth}{{\cellcolor{white}ASC}} & Classify the sentiment of the speaker into one of the following - positivity, negativity, or neutrality & {\textit{neutral}} \\ 
\midrule
\rowcolor{gray!15}
\multirow{1}{0.08\textwidth}{{\cellcolor{white}AC}} & Analyze the speaker voice and figure out their accent from one of American, Indian, Australian, Canadian or British. & {\textit{American}} \\ 
\midrule
\multirow{1}{0.08\textwidth}{{\cellcolor{white}SC}} & How many distinct speakers are there in this audio clip. & {\textit{one}} \\ 
\midrule
\rowcolor{gray!15}
\multirow{1}{0.08\textwidth}{{\cellcolor{white}SNS}} & Is their speech in the preceding audio or not? & {\textit{speech}} \\ 
\bottomrule
\end{tabular}
\egroup
}
\vspace{-0.5cm}
\end{table*}

%% file: 200.background.tex
\section{Related Work}
\label{sec:related_work}

\noindent\textbf{Multi-task learning.} Prior studies have shown that a single deep learning model is capable of jointly learning multiple large-scale tasks across different domains \cite{kaiser2017one}. The key idea in multi-task learning is to leverage shared representations across related tasks to improve overall generalization and efficiency. Following this approach, the T5 model \cite{raffel2020exploring} frames all text tasks as text-in to text-out, using a unified text-based framework that facilitates shared representations across textual tasks. Similarly, SpeechNet \cite{chen2021speechnet} and SpeechT5 \cite{ao2021speecht5} leverage a shared encoder-decoder framework to jointly model speech and text modalities spanning 5 to 6 tasks like TTS, ASR, and Voice Conversion (VC). VIOLA \cite{wang2023viola}, a single auto-regressive Transformer decoder-only network, unifies various cross-modal speech and text tasks as a conditional codec language model via multi-task learning. Whisper \cite{radford2023robust} also employs large-scale multi-task learning, training on related speech tasks including language identification, speech recognition, and translation. In this work, SpeechVerse utilizes multi-task training to transfer knowledge between several related tasks while using natural language instructions to perform each task. Unlike prior work that generated text, speech, or both, our method focuses solely on producing textual output, while taking in audio and text instructions.

% \vspace{2mm}
\noindent\textbf{Multimodal Large Language Models.} Prior work on multimodal LLMs has focused primarily on tasks involving images, such as image generation, visual question answering, and image captioning \cite{achiam2023gpt, alayrac2022flamingo, li2022blip, team2023gemini}. Multimodal models incorporating modalities like audio and speech have received relatively less attention compared to vision-and-language models \cite{zhou2022learning, koh2023grounding, peng2023kosmos}. However, there has been growing interest in augmenting large language models with audio data, leading to several proposed approaches \cite{huang2023audiogpt, zhang2023speechgpt, gong2023listen, rubenstein2023audiopalm, shu2023llasm, deshmukh2023pengi, wang2023slm}. SpeechGPT \cite{zhang2023speechgpt} proposed a multimodal LLM combining discrete units of HuBERT with an LLM to solve few understanding tasks like ASR, Spoken QA as well as generation tasks like TTS. \cite{wang2023slm} introduces the novel capability of zero-shot instruction-following for more diverse tasks such as dialog generation, speech continuation and Question Answering. Most recently, \cite{chu2023qwen} proposed Qwen-Audio, a large-scale audio-language model trained using a multi-task learning approach to handle a diverse range of tasks across various audio types including human speech, natural sounds, music, and songs. Qwen-Audio employs a single audio encoder to process various types of audio whose initialization is based on the Whisper-large-v2 model \cite{radford2023robust} and performs full finetuning. In contrast, our work utilizes two frozen pretrained models, one each for speech encoder and text decoder to retain their intrinsic strengths. Also, we utilize 30+ instructions for each task during training for improved generalization whereas \cite{wang2023slm} uses a single fixed instruction. Additionally, SpeechVerse incorporates multi-task learning and instruction finetuning in a single training stage.

%% file: 500.conclusion.tex
\section{Conclusion}
\label{sec:conclusion}

In this work, we propose SpeechVerse, a multimodal framework that enables LLMs to follow natural language instructions for performing diverse speech processing tasks. 
Through supervised instruction finetuning and combining representations from frozen pre-trained speech and text foundation models, SpeechVerse achieves strong zero-shot generalization on unseen tasks. Extensive benchmarking against conventional baselines show SpeechVerse's superiority on 9 out of 11 tasks, demonstrating its formidable instruction following capability. 
Crucially, SpeechVerse maintains robust performance on out-of-domain datasets, unseen prompts, and even unseen tasks. 
This highlights the efficacy of our proposed training methodology in imbuing the model with a generalizable skill for mapping text-based instructions to speech processing outputs. Moving forward, we aim to expand SpeechVerse's capabilities to follow even more complex instructions and generalize to new domains. 
% We are excited about the potential for instruction-driven systems like SpeechVerse to simplify and democratize speech and audio processing. 
By separating task specification from model design, SpeechVerse represents a versatile framework that can dynamically adapt to new tasks through natural language without retraining.

%% file: 600.ethics.tex
\section*{Limitations}
\label{sec:limitations}

While this work demonstrated strong instruction following capabilities for the multitask SpeechVerse model across a variety of tasks, some limitations remain. The study relied on a single underlying LLM architecture (FlanT5) rather than exploring more recent models tailored for instruction following. Additionally, there is a trade-off between generalized capabilities on unseen tasks versus specialized performance on original training tasks that poses challenges for a single multitask model. While the model showed promise in handling diverse unseen tasks, its limitations were not fully characterized across the wide scope of possible instructions and the performance on these unseen tasks is not quantitatively measured. 

\section*{Ethics Statement}
\label{sec:ethics}

All speech datasets we use have anonymous speakers. We do not have any access to nor try to create any PII (Personal Identifiable Information) of speakers, and our model neither identifies speakers nor uses speaker embeddings. Most of the work used public open-source datasets for both training and testing. The in-house datasets used for pre-training Best-RQ encoder and SNS task are collected via third-party speech data vendors. No additional data collections made concerning to the work carried in this paper.

%% file: a00.example.tex
\section{Appendix}
\label{sec:appendix}

\subsection{Audio Encoder Pre-training}
\label{sec:app-audio-encoder}

Our audio encoder is a 24-layer Conformer model with feature dimension of 768 and attention head of 8. The total number of parameters of this encoder model is 300M. We adopt the BEST-RQ~\cite{chiu2022self} method, which pre-trains the model to predict the masked speech signals with labels generated from a random-projection quantizer. The quantizer projects the speech inputs with a randomly initialized matrix, and performs a nearest-neighbor lookup in a randomly-initialized codebook. Neither the projection matrix nor the codebook is updated during pre-training. We build an internal pre-training dataset containing 300K hours English audios. The pre-training uses mask span of 10 with total effective masking ratio about 40\%. The learning rate schedule follows the transformer learning rate schedule with peak value of 0.0005 and warm-up of 50K steps. AdamW optimizer is adopted with weight decay of 0.01. Since the encoder has 4 times temporal-dimension reduction, the quantization with random projections stacks every 4 frames for projections. We use 16 individual codeboooks, where the vocab size of each codebook is 8192 and the dimension is 16. The model is pre-trained for 500K steps in total.

\subsection{Hyper-parameters}
\label{sec:app-hparam}
We train all our models on a cluster of machines with 8 A100 GPUs, each having 40GB of memory. Pytorch Lightning framework\footnote{\tiny \url{https://www.pytorchlightning.ai}} is used for our implementation. A 5\% subset of the complete training dataset for a model is used as a validation set to choose the important hyper-parameters. For the multi-task models, the sample weights for each dataset are also chosen using this validation set. Further, all models are trained till the validation loss converges and it does not improve for 5 consecutive epochs. The details of the learning rate, warmup steps, batch size for the Multitask-WLM and Multitask-BRQ are summarized in the Table \ref{tab:hrapam}. To adhere to memory constraints of the GPUs, we filter out any training sample where the sequence length of the audio is greater than 900 or the target label is greater than 600. Since, the WavLM Large samples audio features at 50Hz rate, we use two successive 1-D convolution blocks (kernel size=3, stride=2) for the Task-FT and Multitask-WLM model to downsample the audio four times and achieve the desired sampling rate of 12.5Hz. For BestRQ-based audio encoder, the sampling rate is 25Hz and hence the stride of the second convolution block is set to 1 to ensure the output sampling rate is 12.5Hz.

\begin{table}[t]
\caption{Various hyper-parameters for our models}
\label{tab:hrapam}
% \vspace{-2mm}
\centering
\footnotesize
\resizebox{0.49\textwidth}{!}{
\begin{tabular}{lcc}
\toprule
\textbf{Parameter} & \textbf{Multitask-WLM} & \textbf{Multitask-BRQ} \\
\midrule
Audio encoder & WavLM-Large & Best-RQ \\
LLM & Flan-T5-XL & Flan-T5-XL \\
Convolution blocks & 2 & 2\\
Kernel sizes & [3, 3] & [3, 3] \\
Strides & [2, 2] & [2, 1] \\
Audio encoder ampling rate & 50Hz & 25Hz \\
Down-sampling factor & 4x & 2x \\
\# parameters & 2.9B & 3.2B\\
\# trainable parameters & 28.3M & 26.8M\\
LoRA rank(r) & 16 & 16 \\
Learning rate & 0.001 & 0.005 \\
Warm-up steps & 100 & 50000 \\
Effective batch size & 768 & 2048 \\
\bottomrule
\end{tabular}
}
\end{table}

\subsection{Tasks}
\label{sec:app-task}
We provide the details about our training tasks below as well as provide some qualitative examples in the Table \ref{tab:training-task-qual} to better understand the tasks. 

\noindent \textbf{ASR}: We use a combination of 5 publicly available datasets for the ASR task, which totals to 3k hours of paired audio and text data. We evaluate performance on the standard benchmarks for ASR. 

\noindent\textbf{ST}: We train our models to predict translations in multiple different languages from the audios recorded with English speech. The tokenizer of the backbone LLM limits the choice of what can be a potential target language. For our case, we train and evaluate on German, French, and Romanian translations from the EuroParl dataset \cite{jairsan2020a}. We also augment the training data with German and Catalan translations from the CoVost2 \cite{wang2020covost} dataset . 

\noindent \textbf{IC/SF}: We train and evaluate our models on a subset of the SLURP dataset~\cite{bastianelli2020slurp} that consists of 10 intent classes and 4 slot labels. This also allows us to study the generalization ability of our models to unseen class labels and we separately study it in the Section \ref{subsec:generalization}. The intent classes and slot labels that are chosen for the "seen" subset are the ones that occur most frequently in the training data. 
The training prompt used for this task is designed to contain the description of each class label. 

\noindent \textbf{KWE}: The goal of this task is to identify important keywords in the content of the speech in the audio. Since no publicly available dataset exists for this task, we synthetically extract keywords from the ground truth transcripts using a text-based keyword extraction model\footnote{\tiny \url{https://huggingface.co/Voicelab/vlt5-base-keywords}}. These are then used as labels for training and evaluating our models. 

\noindent \textbf{KWS}: This is a binary classification task to detect whether a specified keyword was spoken in the audio or not. We create positive samples by randomly selecting keywords from the ground truth transcripts and negative samples by choosing a keyword that does not appear in the transcript. Positive and negative examples are created in 70-30 ratio respectively for both training and evaluation. 

\noindent \textbf{ER}: For emotion recognition, we classify speech into one of four main emotion classes: neutral, happy, sad, and angry, chosen based on the availability of the training samples in the MSP-Podcast v1.11 dataset \cite{Lotfian_2019_3}. We report metrics on the corresponding four-emotion subset of the Test1 split of the dataset.
% We train on the subset of the MSP-Podcast v1.11 dataset \cite{Lotfian_2019_3} containing these four emotions, upsampling the underrepresented classes. and evaluate on corresponding four-emotion subset of the Test1 split. %\pra{There are more labels in this dataset, why did we pick only 4?} -> david: added reasons about the data availability

\noindent \textbf{ASC}: For audio sentiment classification, we classify speech as positive, negative, or neutral in sentiment. The sentiment labels were obtained by thresholding the valence scale (annotated from 1 to 7) with 3 and 5. We train on the entire training split of the MSP-Podcast v1.11 dataset, and evaluate on the corresponding Test1 split. %\pra{Can we rather use Test Set 1, 2, 3 as shown here https://ecs.utdallas.edu/research/researchlabs/msp-lab/MSP-Podcast.html} -> david: yes, we evaluated on test set 1, which is a common choice.

\noindent \textbf{SC}: For speaker counting, we identify whether one or two speakers are present. We train on segments from Fisher dataset transcripts \cite{Fisher1, Fisher2} with one or two speakers, and evaluate on the Fisher test split used in \cite{paturi23_interspeech}.

\noindent \textbf{AC}: We train our models to classify speech into five accents of English language: Canadian, Indian, Australian, British, and American, using metadata from the Mozilla Common Voice dataset.

\noindent \textbf{SNS}: In this task, we identify whether speech is present in the audio. We collect a diverse set of audios with and without speech for training our models and evaluate them on a combination of  speech segments from Hub5 \cite{HUB5} dataset and held-out non-speech segments in our in-house collection.